\documentclass{article}
\usepackage{spconf,amsmath,graphicx }
\usepackage{amssymb}
\usepackage{bbm}
\usepackage{cite}

\title{DPNet: dual-path network for efficient object detection with lightweight self-attention }
\name{Huimin Shi$^{1}$, Quan Zhou$^{1, *}$\thanks{Corresponding author: Quan Zhou, quan.zhou@njupt.edu.cn,This work is partly supported by NSFC (No. 61876093), NSFJS (No. BK20181393), and NSF (No. IIS-1302164).},Yinghao Ni$^{1}$, Xiaofu Wu$^{1}$, and Longin Jan Latecki$^{2}$
}

\address{$^{1}$National Engineering Research Center of Communications and Networking, \\ Nanjing University of Posts \& Telecommunications, P.R. China.\\
	$^{2}$Department of Computer and Information Sciences, Temple University, Philadelphia, USA.\\}
\begin{document}
%
\maketitle
\begin{abstract}
Object detection often costs a considerable amount of computation to get satisfied performance, which is unfriendly to be deployed in edge devices. To address the trade-off between computational cost and detection accuracy, this paper presents a dual path network, named DPNet, for efficient object detection with lightweight self-attention. In backbone, a single input/output lightweight self-attention module (LSAM) is designed to encode global interactions between different positions. LSAM is also extended into a multiple-inputs version in feature pyramid network (FPN), which is employed to capture cross-resolution dependencies in two paths. Extensive experiments on the COCO dataset demonstrate that our method achieves state-of-the-art detection results. More specifically, DPNet obtains 29.0\% AP on COCO test-dev, with only 1.14 GFLOPs and 2.27M model size for a 320 $\times$ 320 image.
\end{abstract}
\begin{keywords}
Lightweight network, object detection, attention mechanism, convolution neural network
\end{keywords}
\section{Introduction}
\label{sec:intro}
\begin{figure}[ht] 
\centering 
\includegraphics[width=0.43\textwidth]{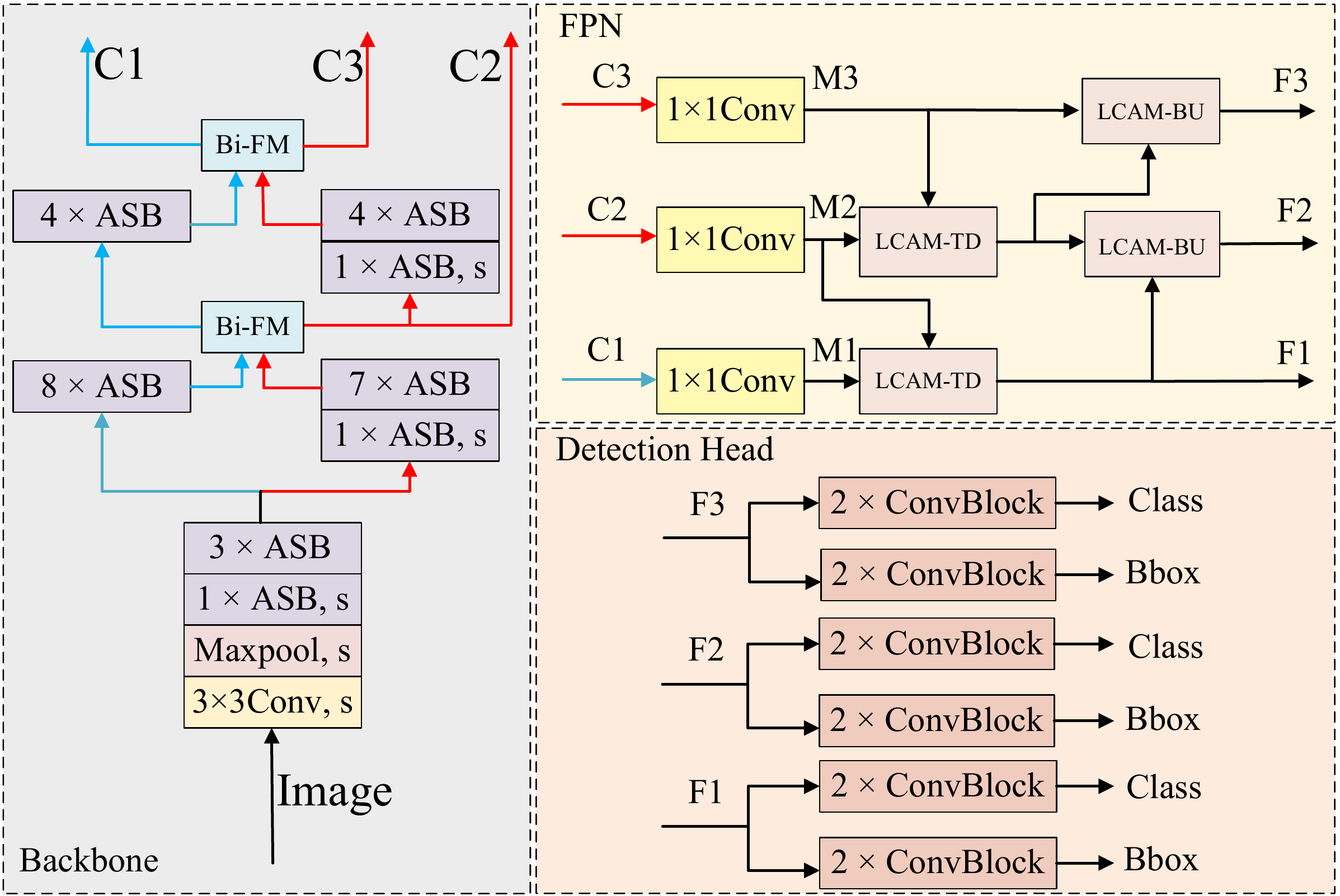} 
\vspace{-0.4cm}
\caption{Overview of DPNet. Our backbone includes a stem and a set of ASBs, resulting in parallel structure of low-resolution path and high-resolution path (denoted by red and blue arrows). Note two paths share stem and first four ASBs. LCAM is designed to enhance cross-scale representations in FPN. Our detection head uses several lightweight ConvBlock units for final prediction. (Best viewed in color)} 
\label{Fig:network} 
\vspace{-0.4cm}
\end{figure}

Object detection is a challenging task in the area of computer vision, which is widely used in self-driving and robot vision. It aims to detect the minimum bounding boxes that cover objects of interest in input image, and assign an associated semantic label synchronously. Recently, the performance of object detectors has been improved tremendously with the advance of convolution neural networks (CNNs). Typically, object detection can be divided into two-stage and one-stage detectors. Two-stage detectors \cite{ren2015faster,he2017mask} are always not efficient due to their multi-stage nature. On the contrary, one-stage detectors \cite{liu2016ssd,redmon2016you} directly predict object categories and bounding box offsets. Despite of achieving various trade-offs between detection accuracy and running speed, their model size and implementing efficiency are still unacceptable for resource-limited equipments (e.g., drones, robots, and smartphones).

Recently, single path lightweight architectures have been dominant for object detection, which can be roughly classified into two categories: a manual design of compact network and neural architecture search (NAS). Motivated from \cite{howard2017mobilenets, sandler2018mobilenetv2, ma2018shufflenet} used for image classification, the first category directly designs lightweight networks in single path manner. For instance, MobileNet-SSD \cite{howard2017mobilenets,sandler2018mobilenetv2} combines MobileNet with SSD-head. ThunerNet \cite{qin2019thundernet} adopts ShufflenetV2 \cite{ma2018shufflenet} as backbone by replacing $3 \times 3$ depth-wise convolution with $5 \times 5$ depth-wise convolution. Pelee \cite{NIPS2018_7466} employs lightweight backbone with dense structure, reducing output scales of SSD-head to save computation. Tiny-DSOD \cite{li2018tiny} introduces depth-wise convolutions both in backbone and feature pyramid network (FPN). Tiny-Yolo families \cite{redmon2017yolo9000,redmon2018yolov3,bochkovskiy2020yolov4} reduce the number of convolution layers or remove multi-scale outputs in FPN. The second category, on the other hand, builds networks using pre-defined network units and searching strategies. For example, MnasNet \cite{tan2019mnasnet} searches models by optimizing accuracy and latency. EfficientNet \cite{tan2019efficientnet} presents a scaling factor to control the balance between network depth, width, and resolution. Mobiledets \cite{xiong2021mobiledets} optimizes latency on different platforms. 

Due to powerful ability for capturing long-range interactions, self-attention \cite{wang2018non,bello2019attention} starts to show its power for object detection. In \cite{hu2018squeeze,hu2018genet}, some efficient attention mechanisms are embedded in compact networks by assigning different weights among feature channels. In spite of achieving remarkable progress, the previous networks inherently suffer from following shortcomings: (1) Single path architecture adopts aggressive downsampling strategies, discarding much fine image details that are useful to locate objects. (2) Although self-attention \cite{wang2018non,bello2019attention} is able to encode global context, matrix product requires huge computations. On the other hand, global pooling is used in \cite{hu2018squeeze,hu2018genet} to extract global context, yet it is weaken in modeling global spatial interactions.

\begin{figure}[t!] 
\centering 
\includegraphics[width=0.43\textwidth]{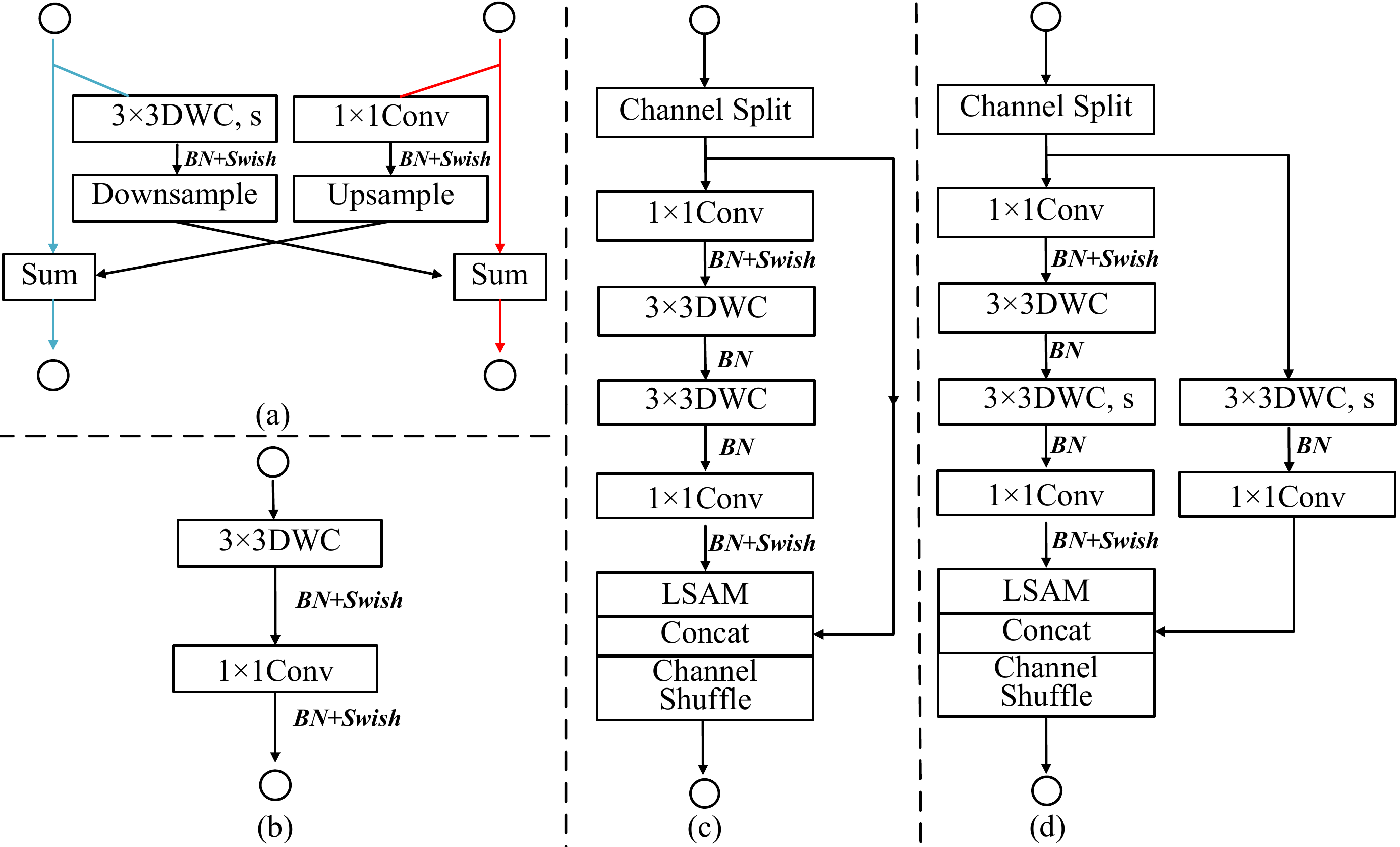} 
\vspace{-0.4cm}
\caption{Overview of the blocks used in backbone and detection head. (a)Bi-FM; (b) ConvBlock; (c) ASB; and (d) stride version of ASB (s = 2). (Best viewed in color)} 
\label{Fig:Blocks} 
\vspace{-0.4cm}
\end{figure}

To address these limitations, this paper describes a light-weight parallel path network, called DPNet, for object detection. As shown in Fig. \ref{Fig:network}, the parallel path architecture leads to dual-resolution subnetwork, extracting high-level semantics in low-resolution path (LRP), and remaining low-level spatial details in high-resolution path (HRP), which are both significant for object detection. Considering the complementary, a bi-direction fusion module (Bi-FM) is constructed to enhance communications between two paths, facilitating the information flow among variable-resolution features. To reduce model size, an Attention-based Shuffle Block (ASB) is designed to build lightweight backbone, in which a single input/output Lightweight Self-Attention Module (LSAM) is employed to capture global interactions. In FPN, LSAM is extended to multiple inputs and single output version, resulting in Lightweight Cross-Attention Module (LCAM) that explores correlations between cross-resolution features. Experimental results demonstrate that DPNet achieves state-of-the-art performance in terms of an available trade-off between detection accuracy and implementing efficiency.

\section{DPNet}

\begin{figure}[ht] 
\centering 
\includegraphics[width=0.375\textwidth]{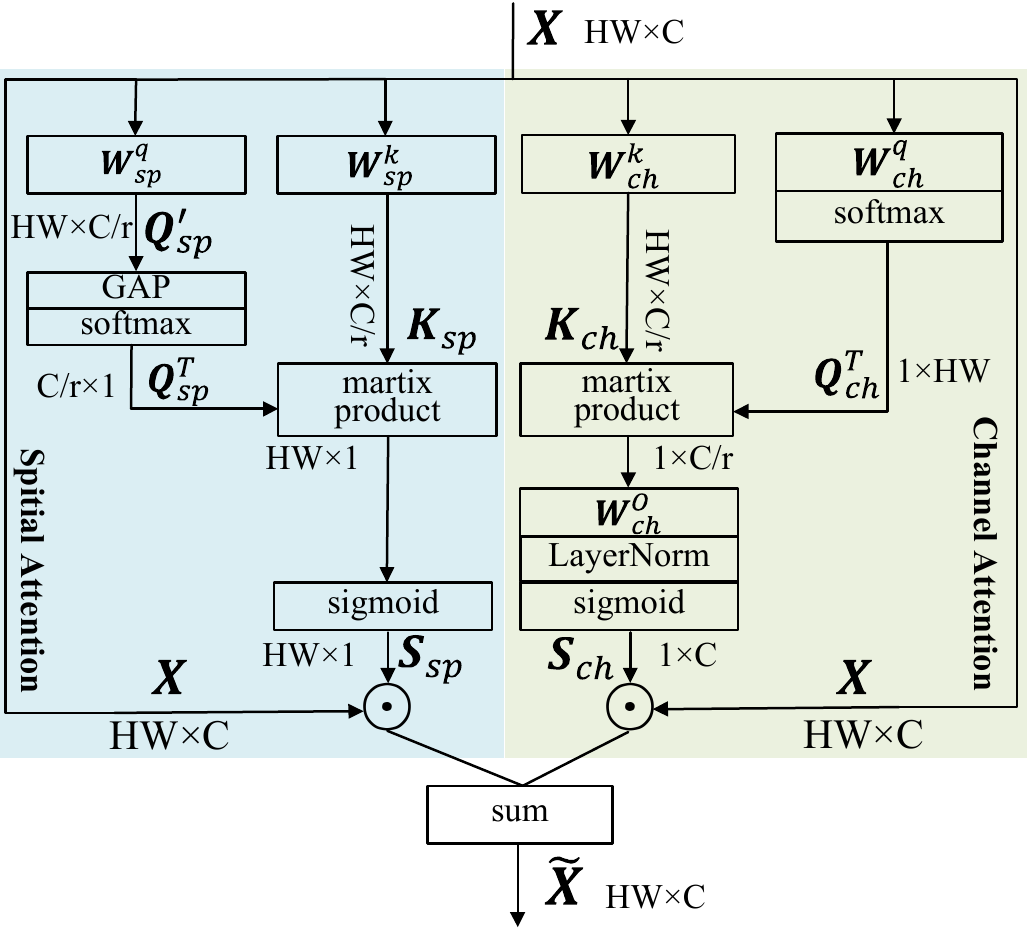} 
\vspace{-0.4cm}
\caption{Overview of LSAM. It includes channel attention and spatial attention. (Best viewed in color)} 
\vspace{-0.35cm}
\label{Fig:LSAM} 
\end{figure}
\label{sec:method}

\subsection{Backbone}
\label{ssec:Backbones}
The backbone of DPNet is mainly constructed by ASB, which is designd to enhance representation ability of global interactions. As depicted in Fig. \ref{Fig:Blocks}(c), at the beginning of each ASB, input features are split into two lower-dimensional branches, where each one has half channels of the input. One branch remains as identity. Another branch serves as residual branch, following the lightweight bottleneck architecture \cite{howard2017mobilenets} to extract features. But the channel numbers in this branch remain unchanged, and the standard convolution is replaced by two depth-wise convolution layers. To capture global context, LSAM is designed and its output is concatenated with identity branch. Finally, feature channels are shuffled to enable information communication between two split branches. After the shuffle, the next ASB unit begins. Fig. \ref{Fig:Blocks}(d) exhibits the stride version of ASB used to reduce feature resolutions, where the $3 \times 3$ stride depth-wise convolutions are utilized in identity branch and second convoluiton layer in residual branch.
{\bf\\LSAM. } Given input feature $\textbf{\emph{F}}$ $\in\mathbbm{R}^{C\times H \times W}$, we first reshape it into a 2D sequence $\textbf{\emph{X}}$ $\in \mathbbm{R}^{HW \times C}$, which is used to compute spatial attention map $\textbf{\emph{S}}_{sp}$ and channel attention map $\textbf{\emph{S}}_{ch}$.

In spatial attention, input sequence $\textbf{\emph{X}}$ is firstly fed into two linear projections $\left\{\textbf{\emph{W}}_{sp}^q, \textbf{\emph{W}}_{sp}^k\right\}\in \mathbb{R}^{C \times C/r}$ to produce low dimensional embeddings $\{\textbf{\emph{Q}}_{sp}^{\prime}, \textbf{\emph{K}}_{sp}\} \in \mathbb{R}^{HW \times C/r}$, where $r$ is a non-negative scale factor. Thereafter, $\textbf{\emph{Q}}_{sp}^{\prime}$ passes a global average pooling $P_g(\cdot)$ to generate $\textbf{\emph{Q}}_{sp}$ $\in \mathbb{R}^{1 \times C/r}$:
\vspace{-0.5em}
\begin{small}
\begin{equation}
\label{eq2}
\begin{aligned} 
    \textbf{\emph{Q}}_{sp}  &= \phi(P_g(\textbf{\emph{X}} \textbf{\emph{W}}_{sp}^q)), \hspace{1cm}                    \textbf{\emph{K}}_{sp}= \textbf{\emph{X}}\textbf{\emph{W}}_{sp}^k
\end{aligned}
\end{equation}
\end{small}
\vspace{-1.5em}
\\where $\phi(\cdot)$ is a softmax function. After that, we perform a matrix product between $\textbf{\emph{K}}_{sp}$ and $\textbf{\emph{Q}}_{sp}^{\mathrm{T}}$ to calculate spatial correlation map, which undergoes a sigmoid function $\sigma(\cdot)$ to generate spatial attention map $\textbf{\emph{S}}_{sp}$ $\in \mathbbm{R}^{HW \times 1}$:
\vspace{-0.5em}
\begin{small}
\begin{equation}
\label{eq2}
\begin{aligned} 
    \textbf{\emph{S}}_{sp}  &= \sigma(\textbf{\emph{K}}_{sp} \textbf{\emph{Q}}_{sp}^{\mathrm{T}})
\end{aligned}
\vspace{-0.8em}
\end{equation}
\end{small}


In channel attention, given two linear projections $\textbf{\emph{W}}_{ch}^q \in \mathbb{R}^{C \times 1}$ and  $\textbf{\emph{W}}_{ch}^k \in \mathbb{R}^{C \times C/r}$, features $\textbf{\emph{Q}}_{ch}$ $\in \mathbb{R}^{HW \times 1}$ and $\textbf{\emph{K}}_{ch}$ $\in \mathbb{R}^{HW \times C/r}$ are similarly generated with $\textbf{\emph{Q}}_{sp}$ and $\textbf{\emph{K}}_{sp}$:
\vspace{-0.5em}
\begin{small}
\begin{equation}
\label{eq2}
\begin{aligned} 
    \textbf{\emph{Q}}_{ch}  &= \phi(\textbf{\emph{X}}\textbf{\emph{W}}_{ch}^q), \hspace{1cm}                    \textbf{\emph{K}}_{ch}= \textbf{\emph{X}}\textbf{\emph{W}}_{ch}^k
\end{aligned}
\vspace{-0.6em}
\end{equation}
\end{small}
Then, channel correlation map is obtained by matrix product between $\textbf{\emph{Q}}_{ch}^{\mathrm{T}}$ and $\textbf{\emph{K}}_{ch}$, which is sequentially fed into a linear projection $\textbf{\emph{W}}_{ch}^O$ $\in \mathbb{R}^{C/r \times C}$, layernorm $LN(\cdot)$, and a sigmoid function to get channel attention map $\textbf{\emph{S}}_{ch}$ $\in \mathbb{R}^{HW \times C}$:
\vspace{-0.5em}
\begin{small}
\begin{equation}\label{eq:LSAM_CA}
\begin{aligned} 
    \textbf{\emph{S}}_{ch}  &= \sigma(LN(\textbf{\emph{Q}}_{ch}^{\mathrm{T}} \textbf{\emph{K}}_{ch}\textbf{\emph{W}}_{ch}^O))
\end{aligned}
\vspace{-0.7em}
\end{equation}
\end{small}
Finally, $\textbf{\emph{S}}_{ch}$ and $\textbf{\emph{S}}_{sp}$ are used to re-weight input $\textbf{\emph{X}}$:
\vspace{-0.6em}
\begin{small}
\begin{equation}
\label{eq1}
\begin{aligned} 
    \widetilde{\textbf{\emph{X}}} = \textbf{\emph{S}}_{sp}\odot \textbf{\emph{X}} \oplus
    \textbf{\emph{S}}_{ch} \odot \textbf{\emph{X}} 
\end{aligned}
\vspace{-0.8em}
\end{equation}
\end{small}
where $\oplus$ and $\odot$ are element-wise addition and product, respectively. The output sequence $\widetilde{\textbf{\emph{X}}}$ $\in \mathbb{R}^{HW \times C}$ is reshaped to $\widetilde{\textbf{\emph{F}}}$ $\in \mathbb{R}^{H \times W \times C}$, and ready for following concatenation.


LSAM leverages computational efficiency and representation ability. Firstly, matrix dimensions of $\textbf{\emph{Q}}_{sp}$ and $\textbf{\emph{Q}}_{ch}$ are significantly reduced by global pooling and $\textbf{\emph{W}}_{ch}^q$, besides, reduction ratio $r$ is set to 8, saving great computation cost in matrix product. Meanwhile, unlike \cite{hu2018squeeze,hu2018genet} that only utilize global pooling to aggregate global clues, LSAM encodes long range dependency by computing correlation matrix.

\subsection{FPN}
\label{ssec:Attention Ehance Feature Pyramid}
\begin{figure}[t!] 
\centering 
\includegraphics[width=0.45\textwidth]{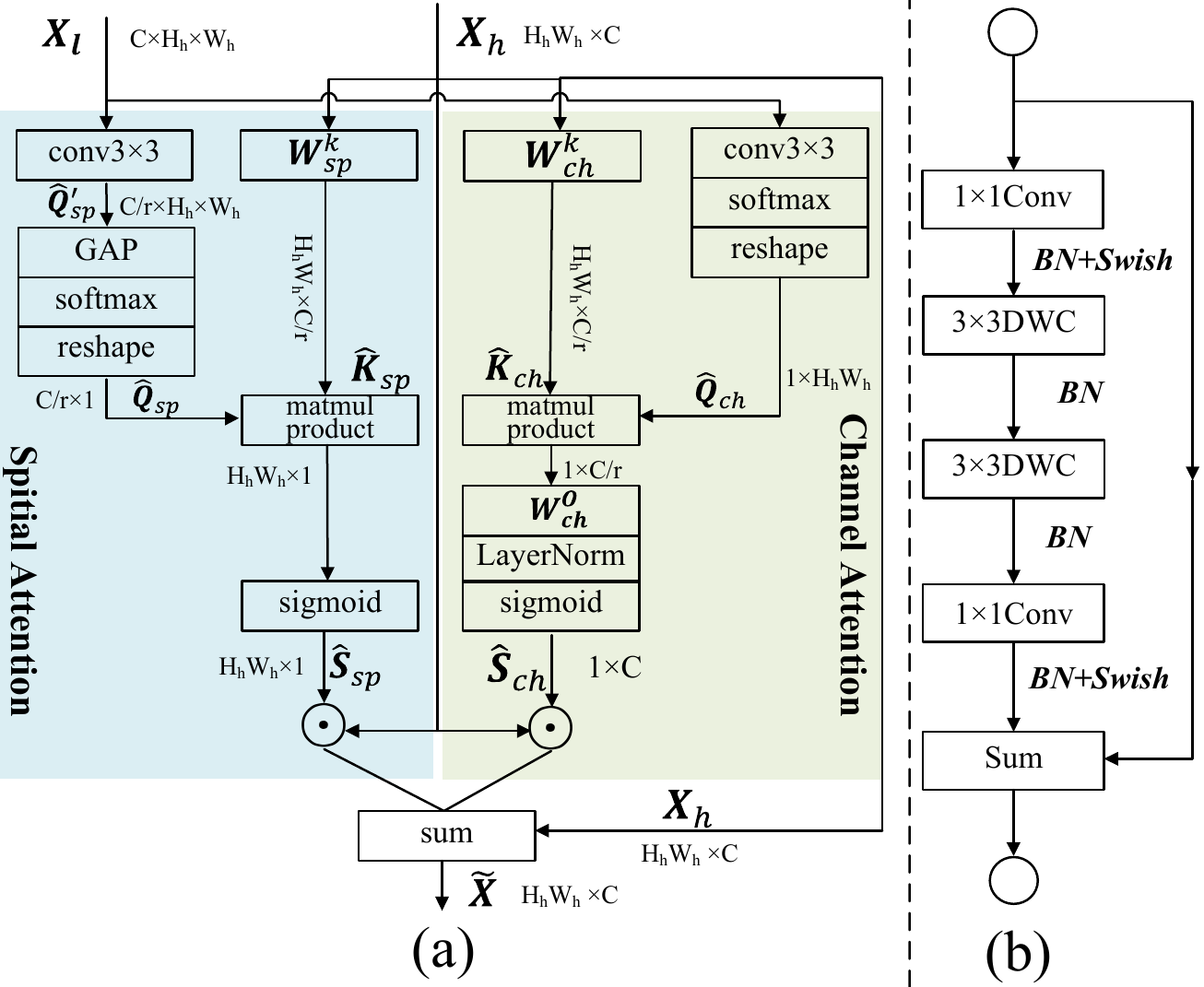} 
\vspace{-0.4cm}
\caption{Overview of the blocks used in FPN. (a) LCAM-TD. (b) Bottleneck. (Best viewed in color) } 
\label{Fig:LCAM} 
\vspace{-0.35cm}
\end{figure}

\begin{table*}
\footnotesize
\vspace{-1.5em}
\caption{Comparison with the state-of-the-art object detectors in terms of detection accuracy and implementing efficiency on COCO test-dev. $AP$ is COCO standard metric, averagely evaluated at IoU ranged from 0.5 to 0.95 with updated step 0.05.} 
\vspace{-1.5em}
	\center
	\begin{tabular}{l|c|c|c|c|c|c|c|c}
		\hline 
		\noalign{\smallskip}
		Method&Year&Backbone&Input Size&FLOPs (G) & Params (M) & $AP$ (\%) &$AP_{50}$ (\%) &$AP_{75}$ (\%)\\
		\noalign{\smallskip}\hline\noalign{\smallskip}
		Tiny-YOLOV3 \cite{redmon2018yolov3} &Arxiv2018&Tiny-DarkNet & $416\times416$&2.78 & 8.7 &16.0&33.1&--\\
		MobileNet-SSD \cite{howard2017mobilenets}  &Arxiv2017 & MobileNet &$300\times300$& 1.2 & 6.8 &19.3&--&--\\
		Tiny-YOLOV4 \cite{bochkovskiy2020yolov4} &Arxiv2020& Tiny-CSPDarkNet &$416\times416$&3.45 & 6.1 &21.7&40.2&--\\ 		
 		MobileNetV2-SSDLite \cite{sandler2018mobilenetv2}  &CVPR2018 & MobileNetV2 &$320\times320$& 0.8 & 4.3 &22.1&--&--\\
 		MobileNet-SSDLite \cite{sandler2018mobilenetv2}  &CVPR2018 & MobileNet &$320\times320$& 1.3 & -- &22.2&--&--\\
 		Pelee \cite{NIPS2018_7466} &NeurIPS2018 &PeleeNet&$304\times304$  &1.29 & 6.0 &22.4&38.3&22.9\\
		Tiny-DSOD \cite{li2018tiny}&BMVC2018 &DDB-Net&$300\times300$  &1.12 & \textbf{1.15} &23.2&40.4&22.8\\
		LightDet \cite{tang2020lightdet}&ICASSP 2020 & LightNet&$320\times320$&\textbf{0.5} & -- &24.0&42.7&24.5\\
		MobileDets \cite{xiong2021mobiledets}&CVPR2021 & IBN+Fused+Tucker  &$320\times320$&1.43 & 4.85 &26.9&--&--\\
		ThunderNet \cite{qin2019thundernet}&ICCV2019&SNet535&$320\times320$&1.3 & -- &28.1&$\textbf{46.2}$&29.6\\
		\noalign{\smallskip}\hline\noalign{\smallskip}
		Ours &- &DPNet&$320\times320$&1.14 & 2.27 &  $\textbf{29.0}$ &45.2&$\textbf{30.4}$\\		
		\noalign{\smallskip}\hline\noalign{\smallskip}
	\end{tabular}
	\label{tab:det_result}
\vspace{-1.5em}
\end{table*}
Feature pyramid is a fundamental component in state-of-the-art detectors, which mainly aggregates multi-scale features via bilinear interpolation and element-wise addition \cite{lin2017feature,liu2018path}. This simply fusion strategy ignores dependencies across features with different size, motivating us to extend LSAM into a multi-inputs version. As depicted in Fig. \ref{Fig:network}, FPN begins at a series of $1 \times 1$ convolutions, producing features of $\lbrace$M1, M2, M3$\rbrace$ with equal channel numbers, yet various resolutions. Thus LCAM is adopted to aggerate cross-resolution features in top-down and bottom-up directions (denoted as LCAM-TD and LCAM-BU). LCAM-TD aims to extract high-level semantics for class identification, while LCAM-BU desires to strengthen low-level details for object localization. As they have similar structure, we only elaborate on LCAM-TD, as shown in Fig. \ref{Fig:LCAM} (a), for convenience understanding.
\\{\bf LCAM-TD.} Let $\textbf{\emph{F}}_h \in \mathbbm{R}^{C\times H_h \times W_h}$ and $\textbf{\emph{F}}_l \in \mathbbm{R}^{C\times H_l \times W_l}$ be high-resolution and low-resolution inputs, we first upsample $\textbf{\emph{F}}_l$ to $\textbf{\emph{X}}_l \in \mathbbm{R}^{C\times H_h \times W_h}$ that matches resolution of $\textbf{\emph{F}}_h$, and reshape $\textbf{\emph{F}}_h$ into a 2D sequence $\textbf{\emph{X}}_h \in \mathbbm{R}^{H_hW_h \times C}$, which are used to compute spatial attention map $\widehat{\textbf{\emph{S}}}_{sp}$ and channel attention map $\widehat{\textbf{\emph{S}}}_{ch}$.


In spatial attention, $\textbf{\emph{X}}_l$ firstly undergoes a $3 \times 3$ convolution $ \textbf{\emph{f}}_{3\times3}^{sp}$ to produce a low dimensional embedding $\widehat{\textbf{\emph{Q}}}_{sp}^{\prime}$ $\in \mathbb{R}^{C/r \times H_h \times W_h}$, where $r$ is a non-negative scale factor. Thereafter,  ${\widehat{\textbf{\emph{Q}}}_{sp}}^{\prime}$ passes global average pooling $P_g(\cdot)$, softmax function $\phi(\cdot)$, and reshape operation step-by-step, generating $\widehat{\textbf{\emph{Q}}}_{sp}$ $\in \mathbb{R}^{C/r \times 1}$. Meanwhile, ${\textbf{\emph{X}}_{h}}$ is fed into a linear projection ${\textbf{\emph{W}}_{sp}^k}$ $\in \mathbb{R}^{C \times C/r}$ to produce ${\widehat{\textbf{\emph{K}}}_{sp}}$ $\in \mathbb{R}^{H_hW_h \times C/r}$:
\vspace{-0.5em}
\begin{small}
	\begin{equation}\label{eq:QK}
	\begin{aligned} 
	\widehat{\textbf{\emph{Q}}}_{sp}  &= \phi(P_g( \textbf{\emph{f}}_{3\times3}^{sp} \ast \textbf{\emph{X}}_l)), \hspace{1cm}                    \widehat{\textbf{\emph{K}}}_{sp}= \textbf{\emph{X}}_{h} \textbf{\emph{W}}_{sp}^k
	\end{aligned}
\vspace{-0.6em}
	\end{equation}
\end{small}
After that, we calculate relation map via matrix product between ${\widehat{\textbf{\emph{Q}}}_{sp}}$ and ${\widehat{\textbf{\emph{K}}}_{sp}}$, which undergoes a sigmoid function $\sigma(\cdot)$ to generate spatial attention map $\widehat{\textbf{\emph{S}}}_{sp}$ $\in \mathbbm{R}^{H_hW_h \times 1}$:
\vspace{-0.5em}
\begin{small}
\begin{equation}
\label{eq2}
\begin{aligned} 
    \widehat{\textbf{\emph{S}}}_{sp}  &= \sigma(\widehat{\textbf{\emph{K}}}_{sp} \widehat{\textbf{\emph{Q}}}_{sp})
\end{aligned}
\vspace{-0.8em}
\end{equation}
\end{small}
In channel attention, given $3 \times 3$ convolution $\textbf{\emph{f}}_{3\times3}^{ch}$ and linear projection $\textbf{\emph{W}}_{ch}^k \in \mathbb{R}^{C \times C/r}$, features $\widehat{\textbf{\emph{Q}}}_{ch}$ $\in \mathbb{R}^{1 \times H_hW_h}$ and $\widehat{\textbf{\emph{K}}}_{ch}$ $\in \mathbb{R}^{H_hW_h \times C/r}$ are similarly produced with $\widehat{\textbf{\emph{Q}}}_{sp}$ and $\widehat{\textbf{\emph{K}}}_{sp}$ defined in Eqn. (\ref{eq:QK}):
\vspace{-0.5em}
\begin{small}
\begin{equation}
\label{eq2}
\begin{aligned} 
    \widehat{\textbf{\emph{Q}}}_{ch}  &= \phi(\textbf{\emph{f}}_{3\times3}^{ch} \ast\textbf{\emph{X}}_l), \hspace{1cm}                    \widehat{\textbf{\emph{K}}}_{ch}= \textbf{\emph{X}}_{h} \textbf{\emph{W}}_{ch}^k
\end{aligned}
\vspace{-0.6em}
\end{equation}
\end{small}
Similar with Eqn. (\ref{eq:LSAM_CA}), channel attention map $ \widehat{\textbf{\emph{S}}}_{ch} \in \mathbbm{R}^{1 \times C}$ is generated by sequentially performing linear projection $\textbf{\emph{W}}_{ch}^O$ $\in \mathbb{R}^{C/r \times C}$, layernorm $LN(\cdot)$, and sigmoid function $\sigma(\cdot)$:
\vspace{-0.5em}
\begin{small}
\begin{equation}
\label{eq2}
\begin{aligned} 
    \widehat{\textbf{\emph{S}}}_{ch}  &= \sigma(LN(\widehat{\textbf{\emph{Q}}}_{ch} \widehat{\textbf{\emph{K}}}_{ch}\textbf{\emph{W}}_{ch}^O)
\end{aligned}
\vspace{-0.6em}
\end{equation}
\end{small}
Finally, $\widehat{\textbf{\emph{S}}}_{ch}$ and $\widehat{\textbf{\emph{S}}}_{sp}$ are used to re-weight high-resolution input $\textbf{\emph{X}}_h$, which serves as the residual function to facilitate training LCAM-TD in an end-to-end manner:
\vspace{-0.7em}
\begin{small}
	\begin{equation}
	\label{eq1}
	\begin{aligned} 
	\widetilde{\textbf{\emph{X}}} =(\widehat{\textbf{\emph{S}}}_{sp}\odot \textbf{\emph{X}}_h \oplus
	\widehat{\textbf{\emph{S}}}_{ch} \odot \textbf{\emph{X}}_h) \oplus \textbf{\emph{X}}_h
	\end{aligned}
\vspace{-0.6em}
	\end{equation}
\end{small}
The output sequence $\widetilde{\textbf{\emph{X}}} \in \mathbb{R}^{H_hW_h \times C}$ is reshaped to $\widetilde{\textbf{\emph{F}}} \in \mathbb{R}^{H_h \times W_h \times C}$, which is fed into a bottleneck module, as shown in Fig. \ref{Fig:LCAM}(b), and ready for forthcoming feature integration.

LCAM-BU works in a similar way with LCAM-TD except two steps: (1) When computing spatial attention, $\textbf{\emph{X}}_h$ has to be downsampled to match resolution of $\textbf{\emph{X}}_l$; (2) $\textbf{\emph{F}}_l$ has to be reshaped into a 2D sequence with dimension of ${H_lW_l \times C}$ for feature re-weighting and identity mapping.


\subsection{Network Architecture}
The overall pipeline of DPNet is shown in Fig. \ref{Fig:network}. We take the image with size of 320 $\times$ 320 as an input. Concretely, DPNet consists of three components: backbone, FPN and detection head. In backbone, we have dual parallel paths: HRP and LRP. HRP keeps a relatively high resolution (1/8 of image size), while LRP employs multi-step downsampling operations to obtain low-resolution features (1/32 of image size). As shown in Fig. \ref{Fig:Blocks}(a), Bi-FM is designed to enhance cross-resolution feature communication. Our backbone has multi-outputs $\lbrace$C1, C2, C3$\rbrace$, whose shapes are $\lbrace$40 $\times$ 40 $\times$ 128, 20 $\times$ 20 $\times$ 256, 10 $\times$ 10 $\times$ 512$\rbrace$. FPN takes $\lbrace$C1, C2, C3$\rbrace$ as inputs, and produces intermediate features $\lbrace$M1, M2, M3$\rbrace$ with equal feature channels of 128. Then, they are fed into a series of LCAMs in top-down and bottom-up manner, producing outputs of FPN $\lbrace$F1, F2, F3$\rbrace$. These fused cross-resolution features pass two lightweight ConvBlocks, as shown in Fig. \ref{Fig:Blocks}(b), to predict object bounding boxes and categorical labels.


\begin{figure}[t!] 
\centering 
\includegraphics[width=0.484\textwidth]{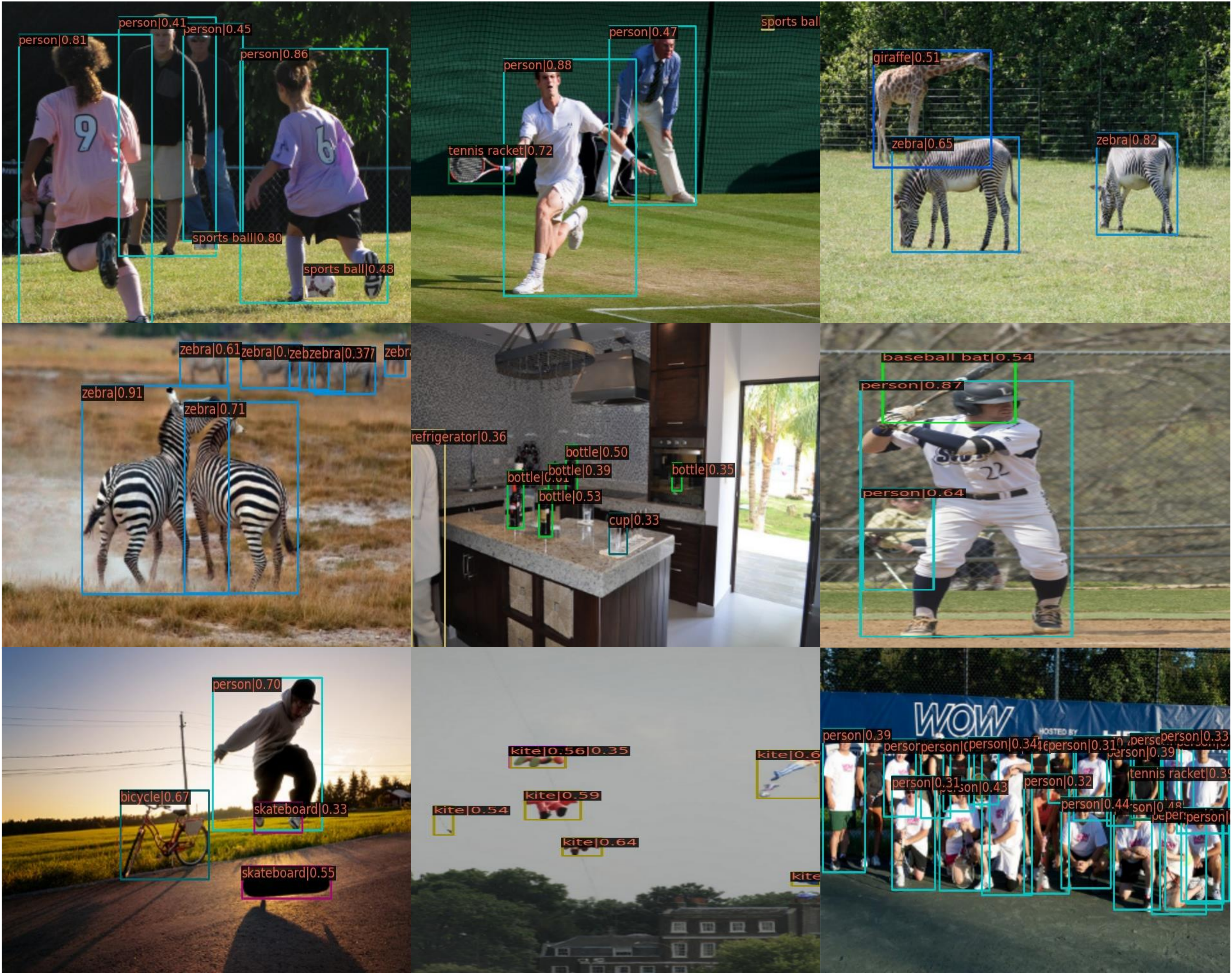} 
\vspace{-0.8cm}
\caption{Visual Examples of qualitative  detection results on COCO test-dev. (Best viewed in color)} 
\label{Fig:Vis} 
\vspace{-0.4cm}
\end{figure}

\section{EXPERIMENTS}
\label{sec:majhead}
\subsection{Experimental Settings}
\label{ssec:subhead}

{\bf Dataset and evaluation protocol.} Experiments are implemented on the widely-used challenging detection benchmark MS COCO 2017 \cite{lin2014microsoft}, which contains 118K training, 5K validation and 20K test-dev images. Our model is trained on the train set. A system-level comparison is reported on test-dev.

\noindent {\bf Implementation Details.} DPNet is trained from scratch with a minibatch of 76 images on a single RTX 2080Ti GPU. We adopt stochastic gradient descent (SGD) for 300 epochs with 5 epochs warm up. Cosine learning strategy is used with an initial learning rate, weight decay, and momentum set as $1.5 \times e^{-2}$, $5 \times e^{-4}$, and $9 \times 10^{-1}$, respectively. We adopt half-precision (FP16) to reduce GPU memory usage. Exponential moving average is adopted to accelerate convergence. CIoU loss \cite{zheng2020distance} is utilized for offset regression of bounding boxes, classification loss and other settings follow \cite{li2020generalized}. Instead of using fancy methods as in \cite{bochkovskiy2020yolov4,zhang2017mixup}, we also follow SSD \cite{liu2016ssd} to augment training data, which is useful to boost performance.

\subsection{Evaluation Results on MS COCO}
\label{ssec:subhead}
Tab. \ref{tab:det_result} reports comparison results with selected state-of-the-art lightweight detectors, demonstrating that DPNet achieves best trade-off in terms of detection accuracy and implementation efficiency. DPNet achieves 29.0\% AP on COCO test-dev, with only 2.27M model size and 1.14 GFLOPs. Among all baselines, DPNet is nearly $2\times$ model size than Tiny-DSOD \cite{li2018tiny}, but improves detection performance with 5.8\% AP. Although LightDet \cite{tang2020lightdet}, another lightweight detectors, has approximately half GFLOPs of DPNet, but delivers poor detection results of 5.0\% drop in terms of AP. Although ThunderNet \cite{qin2019thundernet} outperforms our DPNet in terms of AP$_{50}$, we improve AP and AP$_{75}$ by large margins of 0.9\% and 0.8\%, respectively, indicating that DPNet performs better in object localization. Fig. \ref{Fig:Vis} shows some visual examples of detection results on COCO test-dev. It demonstrates that DPNet accurately detects objects within different scales, such as ``bottle'' in fifth example as well as ``person'' in first example.
\vspace{-0.0cm}

\section{CONCLUSION remarks and future work}
\label{sec:CONCLUSION}

This paper has described a DPNet for lightweight object detection, which extracts high-level semantics and low-level details using dual-resolution representations. In backbone, LSAM is designed in ASB to capture global context, while remains lightweight architecture. We further extend LSAM to LCAM in FPN that encodes global interactions between cross-resolution features. The experimental results on COCO benchmark show that DPNet achieves state-of-the-art trade-off in terms of detection accuracy and implementing efficiency. In the future, we are interested in transferring DPNet to instance segmentation \cite{wang2020solo} and keypoint estimation \cite{sun2019deep}.

\bibliographystyle{ieeetr}
\bibliography{ref.bib}
\end{document}